# Real-Time Sleepiness Detection for Driver State Monitoring System

Deepak Ghimire, Sunghwan Jeong, Sunhong Yoon, Sanghyun Park, Juhwan Choi

IT Application Research Center, Korea Electronics Technology Institute,
Jeonju-si, Jeollabuk-do 561-844, Rep. of Korea
{deepak, shjeong, yoonsh, shpark, netside}keti.re.kr

**Abstract.** Driver face monitoring system can detect driver fatigue, which is an important factor in a large number of accidents, using computer vision techniques. In this paper we present a real-time technique for driver eye state detection. At first face is detected and the eyes are searched inside face region for tracking. A normalized cross correlation based online dynamic template matching technique with combination of Kalman filter tracking is proposed to track the detected eye positions in the subsequent image frames. Support vector machine with histogram of orientation gradient features is used for classification of state of the eyes as open or closed. If the eye(s) state is detected as closed for a specified amount of time the driver is considered to be sleeping and an alarm will be generated.

## 1   Introduction

With the vast research in Intelligent Transportation System (ITS), the improvement of public safety by reducing accidents due to driver fatigue is an important factor. In different countries statistics shows that large amount of accidents occurs due to driver fatigue. In general, the main reason of about 20% of the crashes and 30% of fatal crashes is the driver drowsiness and lack of concentration [1]. In the year 2009, the US National Sleep Foundation (NSF) reported that 54% of adult drivers have driven a vehicle while feeling drowsy and 28% of them actually fell asleep [2]. The German Road Safety Council (DVR) claims that one in four highway traffic fatalities are result of momentary driver drowsiness [3]. According to the study reported in 2007, it is expected that the amount of crashes will be reduced by 10%-20% using driver monitoring systems [4]. Therefore, developing the system for monitoring a driver state and alerting the driver, when he is drowsy and not paying adequate attention to the road, is essential to prevent accidents.

   The possible techniques for detecting drowsiness in drivers can be broadly divided into three major categories [5]: methods based on driver's current state; methods based on driver performance; and methods based on combination of the driver's current state and driver performance. Among them methods based driver's current state uses driver face monitoring, especially driver's eye state monitoring to detect drowsiness.

There has been wide research on detection of fatigue effects and the driver's current state specially focusing on driver's eye state detection such as eye closure, gaze direction, blinking rate etc. A comprehensive survey of research on driver fatigue detection is presented in [5]. Recently, in 2012, a review regarding detecting driver drowsiness based on sensors is presented by Sahayadhas et al. [6]. The driver face monitoring systems can be divided into two general categories. In one category only eye region is processed to detect driver drowsiness whereas in the other category, whole face region is processed to detect driver drowsiness. Processing only eye region instead of processing whole face region has less computational complexity, but as we know symptoms of fatigue not only occur in eye region but also from other regions of face and head such as yawing. Methods presented in [1, 7-11] utilize information only from eye region in order to detect driver's drowsiness whereas methods presented in [12-14] utilize information from not only eye region but also from whole face area in order to detect driver's drowsiness.

In this paper we focused on developing driver drowsiness detection system with less computational overhead. Therefore we only monitor driver's eye state by analyzing eye closure to detect driver drowsiness. The main idea is to monitor eye state for an interval of time and if the eye state is determined closed for a specified amount of time the alarm will be generated to alert the driver. The proposed system works even if single eye state is detected such as in the case of profile view. The advantages of the proposed system over other system presented in the literature is that it can operate in real time with less computational overhead, new tracking scheme is presented so that it can operate in dynamic environment such as change in lighting, works with profile view face, and with the use of support vector machine classification the eye state will be detected accurately even if driver is wearing eyeglass.

## 2    Proposed System

The overall flow chart of the proposed system is shown in Fig. 1. At first the face detection is performed before searching for eyes to reduce the search space as well as false detections. Once the face is detected, eyes are searched inside face region utilizing knowledge regarding face geometry. New eye tracking algorithm is proposed to track the eye positions in the subsequent frames. As soon as tracking is lost the system is re-initialized from face detection step. In each frame, the features are calculated form eye regions to classify eye state as open or closed. Finally the temporal result of the eye state detection is analyzed to decide either the driver is actually drowsy or not. In the following subsection each step of the proposed driver drowsiness detection system are explained in detail.

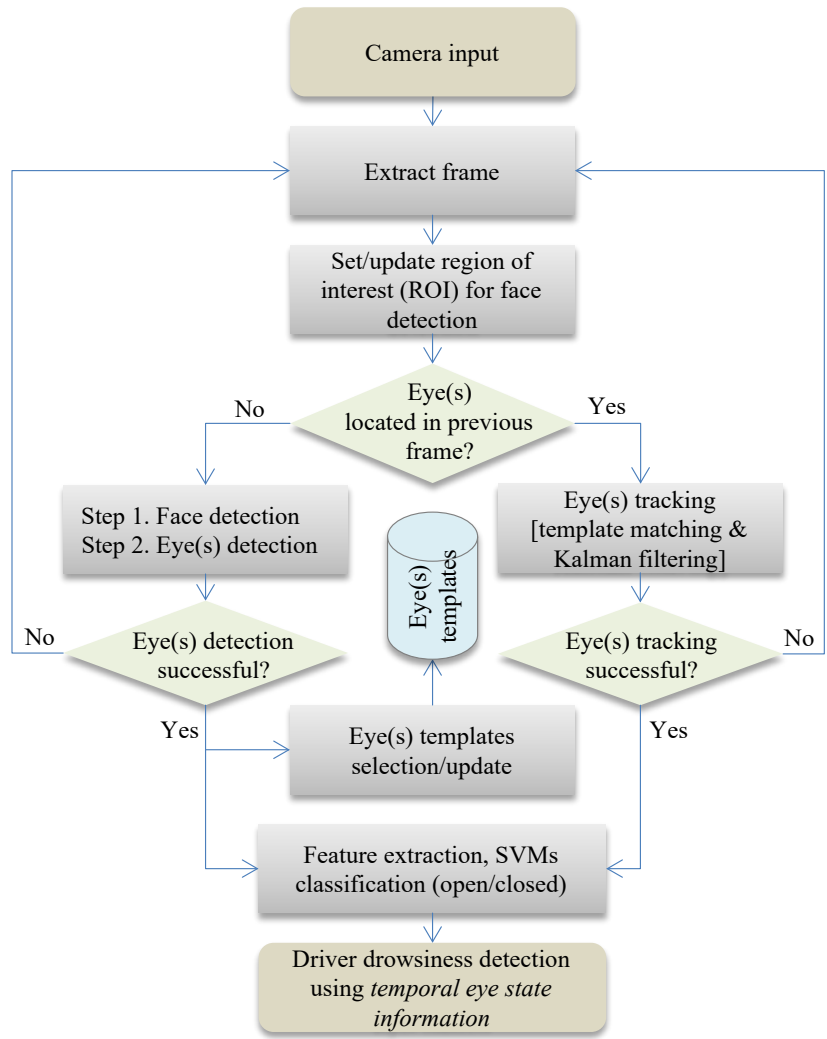

**Fig. 1.** Flowchart of the proposed driver's drowsiness detection system

### 2.1 Face and Eye Detection

Face detection is the primary step in driver drowsiness detection system. Several methods are proposed in the literature for face detection in gray scale images (ex. [15]), as well as in color images (ex. [16]). In our system we used the popular face detection method for gray scale images presented by Viola and Jones [15]. But instead of using Haar like features cascade we used LBP feature cascade as our system is targeted to be implemented in embedded system and LBP cascade is much faster as compared with Haar cascade. Once the face is detected the eyes inside the face region are searched using [15] with Haar cascade. We used OpenCV [17] implementations for both face and eye localization.

## 2.2 Eye Tracking using Dynamic Template Matching

Once the eyes are localized in face for the first time, in subsequent frames, cross correlation based template matching is used for eye tracking. The region of interest (ROI) for searching eyes is set according to previous eye locations. We call our tracking scheme as dynamic template matching based tracking because we collect eye template at the time of eye detection using Haar cascade [15]. Every time if tracking is lost, the eyes are detected using Haar cascade and those detected eyes are stored as template. In tracking time, we use several templates for finding best matching position inside the eye search region based on correlation value. Finally the best matching position is determined by averaging the locations given by several templates. If the matching score is less than specified threshold, eye detection is again performed using face detection followed by eye detection with the help of Viola & Jones technique [15]. The eye template are updated replacing previously stored template so that at any time the stored templates are the most recently detected eye templates using Viola & Jones technique [15]. In the proposed system we used 10 to 20 templates for tracking each eyes.

The following correlation equation is used to find the correlation value between eye template and eye search locations.

$$R(x,y) = \frac{\sum_{x',y'}(T'(x',y') \cdot I'(x+x',y+y'))}{\sqrt{\sum_{x',y'}(T'(x',y'))^2 \cdot \sum_{x',y'}I'(x+x',y+y')^2}} \quad (1)$$

where,

$$T'(x',y') = T(x',y') - 1/(w.h) \cdot \sum_{x'',y''} T(x'',y'')$$
$$I(x+x,y+y) = I'(x+x',y+y') - 1/(w.h) \cdot \sum_{x'',y''} T(x+x'',y+y'') \quad (2)$$

The left and right eye tracking is performed independently and at last the identified left and right eye locations are verified using knowledge regarding eyes geometry inside face region. The final result of eye detection and tracking is subjected to Kamlan tracking for smoothing the tracking result obtained via template matching.

## 2.3 Eye Sleepiness Detection

Once the tracking step is complete, in each frame eye region is subjected for feature extraction and classification as open or closed eye using linear support vector machine classification. Preprocessing is performed in order to improve the classification accuracy. We performed experiment with two different texture features for classification of open or closed eyes; local binary pattern (LBP) [18] and histogram of orientation gradient (HOG) [19] features. The feasibility of each feature will be explained in experimental section in detail. In our system HOG feature outperformed LBP features for eye state classification.

The eye region is subjected for preprocessing before extracting features. Following steps are used for preprocessing: (a) Gamma correction; (b) Difference of Gaussian (DOG) filtering; and (c) Contrast equalization. Fig. 2 shows result of face detection, eye detection, eye region preprocessing, HOG feature extraction and SVM classification as open or closed eyes.

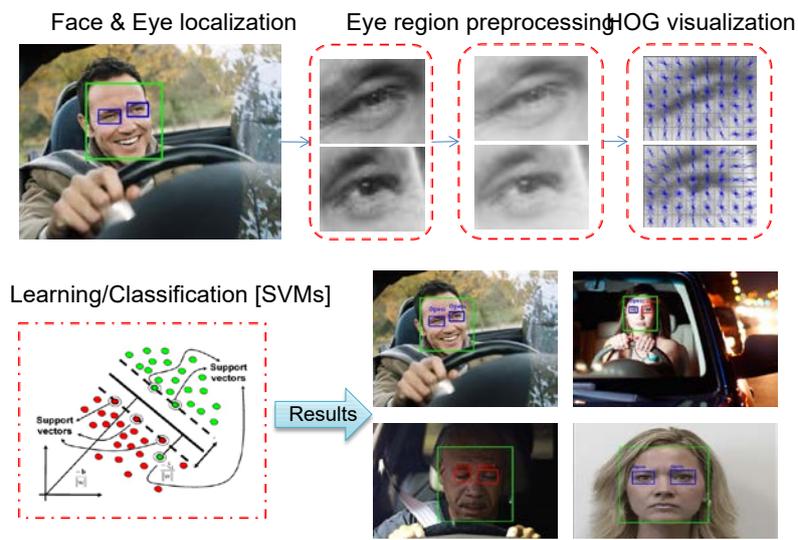

**Fig. 2.** Example of eye state detection using SVM classification with HOG features

## 3   Experimental Results

The proposed driver state detection algorithm was tested on vehicle driving by various persons as well as in lab environment. In our system the camera is not placed directly in front of the driver but in little left position. The main reason was not to distract the driver so that he/she will not feel the camera is monitoring his/her activity. Sometime if driver is turned right, only one eye is properly visible to the camera. In such situation we made decision of diver's current state based on result obtained from single eye state monitoring.

As we used support vector machine for learning driver's eye state as open or closed, eye images from RIP ISL Eye Dataset [20] and personal collection are used for training the SVM classifier. We used 9763 eye images for training in which 4180 are closed and 5583 are open eye images. At first correlation based template matching algorithm is used for tracking driver eyes. Fig. 4 shows some sample image frames of eye tracking. This sample video lasts for 2 min and 59 seconds which consists of 5000 frames. We track both eyes independently to each other, but the final result of tracking is analyzed in term of both eye positions in order to verify the tracking result.

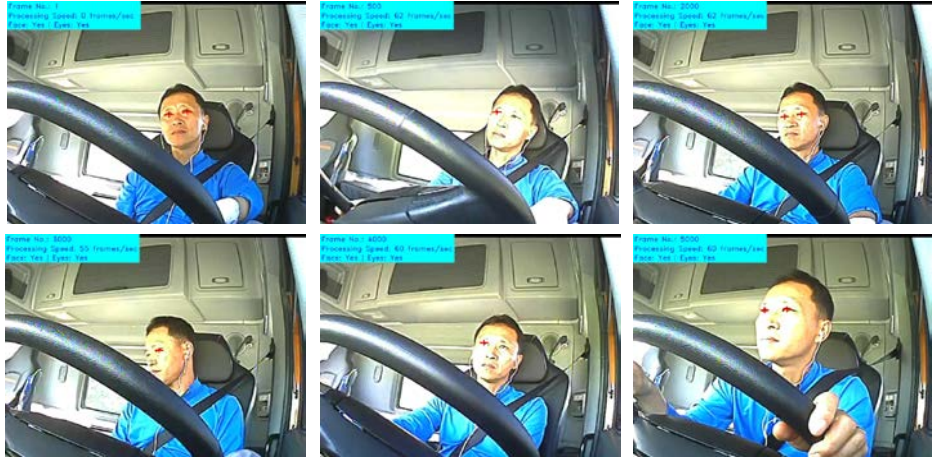

**Fig. 4.** Sample image frames from a video sequence showing tracking result.

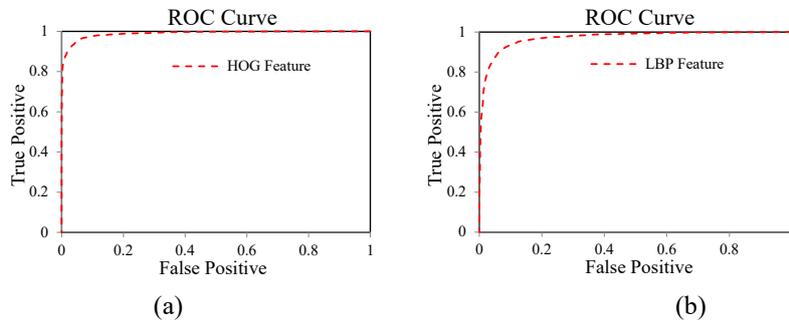

**Fig. 5.** ROC curve of the eye state classification using SVM with (a) LBP and (b) HOG feature descriptor.

LBP and HOG features are studied in order to discriminate open and closed eye with support vector machine classification. Small region around detected eye position as shown in fig. 2 is extracted and features are calculated. In case of LBP feature extraction we divide eye region into 3x2 grids resulting in 348 dimensional LBP feature descriptor whereas in case of HOG feature extraction we divide 48x32 eye region into 8x8 cell size resulting in 540 dimensional HOG feature descriptor. The SVM classification performance for each feature is shown as ROC curve in Fig. 5. The HOG feature outperformed LBP feature for eye state detection in our experiment. Finally, Fig. 6 shows some result of eye sleepless detection in lab environment as well as in vehicle. The system is designed in such a way that even if single eye is detected it will be able to detect the driver state. The result of eye state detection in specified number of frames or time duration is used to decide either the driver is actually sleeping or not which is used for generating alarm to alert the driver.

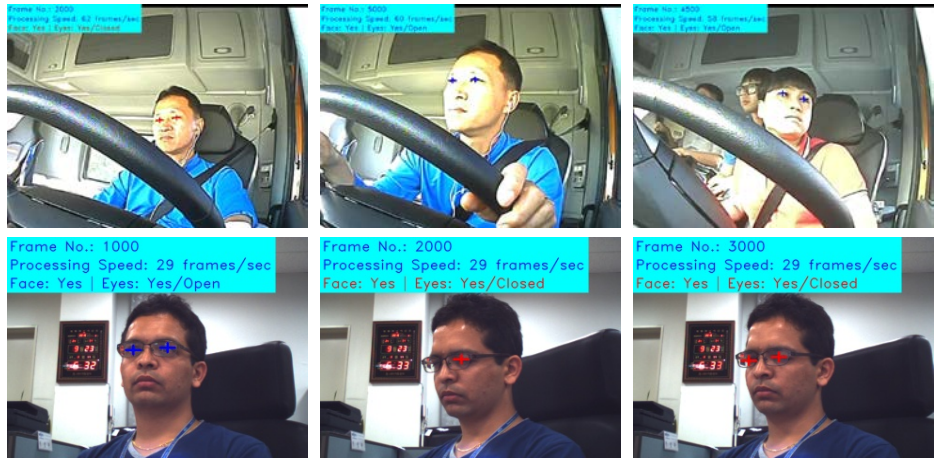

**Fig. 6.** Example results of eye state detection in vehicle as well as in lab environment. All detections are correct.

## 4 Conclusions

In this paper, a new adaptive method for driver drowsiness detection was proposed by monitoring the state of eyes in video sequence. The correlation based template matching was used for tracking the detected eye positions. Two different feature descriptors (LBP and HOG) were studied for learning driver eye state in each frame. Thus obtained eye state result in each frame was utilized for deciding driver sleepiness which was used for generating the alarm to alert the driver.

The proposed method was tested in video sequence recorded in vehicle as well as in lab environment. The proposed system works in real time with minimal computational complexity. Therefore it is also suitable for implementing in embedded environment. The future work will focus on utilizing driver expressions not only from eye regions but also from other face regions to monitor the driver drowsiness with minimal computation overhead.

**Acknowledgements.** This work was supported by the National IT Industry Promotion Agency and funded by Korean Ministry of Science, ICT and Future Planning (Supporting project of Regional SW Convergence Product Commercialization, 2014 ~ 2015).